\documentclass[runningheads]{llncs}
\usepackage{graphicx}
\usepackage{hyperref}
\usepackage{url}
\usepackage{arydshln}
\usepackage{array}
\usepackage{tabularx}
\usepackage{booktabs}
\usepackage{multicol}
\usepackage{multirow}
\usepackage{xcolor}
\usepackage[misc]{ifsym}

\begin{document}
\title{LimGen: Probing the LLMs for Generating Suggestive Limitations of Research Papers}
\titlerunning{LimGen}
\tocauthor{Abdur Rahman Bin Mohammed Faizullah, Ashok Urlana, Rahul Mishra}
\toctitle{LimGen: Probing the LLMs for Generating Suggestive Limitations of Research Papers}

\author{Abdur Rahman Bin Mohammed Faizullah\inst{1, 2, *} \and
Ashok Urlana\inst{3, *} \and \\ Rahul Mishra\inst{2}}
\authorrunning{A.R.B.M. Faizullah et al.}
% If there are two authors, write 'A.L. Benjamin and C.C. Broadus Jr.'
% If there are more than two authors, '[...] et al.' is used.
 \institute{ Infosys Limited, Hyderabad, India \\
 \email{abdurrahman.b@infosys.com} \\ \and
 Language Technologies Research Center, KCIS, IIIT Hyderabad \\
\email{abdur.rahman@research.iiit.ac.in, rahul.mishra@iiit.ac.in}\\ \and
 TCS Research, Hyderabad, India\\
\email{ashok.urlana@tcs.com}}

\maketitle              % typeset the header of the contribution
\def\thefootnote{*}\footnotetext{Equal contribution.}\def\thefootnote{\arabic{footnote}}
\begin{abstract}
Examining limitations is a crucial step in the scholarly research reviewing process, revealing aspects where a study might lack decisiveness or require enhancement. This aids readers in considering broader implications for further research. In this article, we present a novel and challenging task of Suggestive Limitation Generation (SLG) for research papers. We compile a dataset called \textbf{\textit{LimGen}}, encompassing 4068 research papers and their associated limitations from the ACL anthology. We investigate several approaches to harness large language models (LLMs) for producing suggestive limitations, by thoroughly examining the related challenges, practical insights, and potential opportunities. Our LimGen dataset and code can be accessed at \url{https://github.com/arbmf/LimGen}.
% \url{https://anonymous.4open.science/r/LimGen-7834}.

% \url{https://github.com/armbf/LimGen}.

\keywords{Limitations \and LLM \and Constrained Text Generation.}
\end{abstract}

\section{Introduction}
The process of reviewing research articles lies at the core of the scientific publication process. It serves as a pivotal stage wherein experts meticulously scrutinize the content, methodology, and findings of a study before it is disseminated to the broader scientific community. This rigorous evaluation and critique of the strengths and limitations of the research work ensures the integrity, reproducibility, and credibility of the research being presented. Identifying the potential limitations of the research papers is a very tedious and time-consuming process that involves delving into the intricacies of the research methodology, understanding the theoretical framework, and grasping the nuances of the subject matter. This endeavor demands not only a broad knowledge base but also keen attention to detail and analytical thinking.\\  
\begin{table}[t]
    \centering\footnotesize
    \caption{Comparison of actual vs LLM generated suggestive limitations (best viewed in color). \textcolor{magenta}{Magenta} indicates the matching content between the actual and LLM-generated limitations, whereas \textcolor{blue}{Blue} represents the novel limitations.}
    \begin{tabularx}{\linewidth}{X}
        \toprule
        \textbf{Actual Limitation:} Contamination is a significant problem in the era of pretrained language models trained on large web corpora... Additionally, this data's distribution is fixed to pre-college Physics, Chemistry, and Mathematics. There are more gradations and difficulty levels at which the model can be evaluated, which have not been tested as part of our analysis. \\
        \midrule
        \textbf{Llama2-DPR:} The limitations of the approach are: The \textcolor{magenta}{dataset is limited to only three subjects (Physics, Chemistry, and Mathematics)} and \textcolor{blue}{does not cover other important subjects like Biology, Geology, and Computer Science}. The dataset does not include any real-world applications or scenarios, limiting the generalization of the model's capabilities. The \textcolor{magenta}{evaluation metric} used is \textcolor{blue}{based on the number of correct answers}, which does not fully capture the model's reasoning abilities. The \textcolor{blue}{dataset does not include any adversarial examples} or challenging cases to test the model's robustness and adaptability.\\
        \bottomrule
    \end{tabularx}%
    \label{tab:limGen_output}
\end{table}
This paper presents a novel task of Suggestive Limitation Generation (SLG) for research papers, which aims to generate a diverse array of potential limitations specific to each paper, providing reviewers with valuable insights to facilitate their assessment process. The task of SLG poses a greater challenge compared to text summarization, question-answering (QnA), and open-ended text generation tasks. It demands not only reasoning abilities but also the capacity to discern and incorporate associations from a corpus of previously encountered papers during the fine-tuning process. For example, in Table~\ref{tab:limGen_output}, we show the actual limitation block from a research paper in row 1 and the corresponding generated limitations in row 2. Illustrated in magenta, the proposed model demonstrates the capability to generate limitations akin to those originally outlined in the paper (\textit{the dataset is limited to only three subjects, Physics, Chemistry, and Mathematics}) Additionally, it showcases the ability to propose novel, valid limitations (depicted in blue) related to adversarial examples, a facet not included by authors in original limitations.

The key idea behind this approach is to capitalize on cues regarding the similarities or differences among research papers and learn to recommend comparable limitations for a paper, especially when its underlying methodology closely aligns with that of a set of other papers. To this end, we create a dataset called \textbf{\textit{LimGen}}, comprising 4068 research papers and corresponding limitations from the ACL anthology. Subsequently, we probe many state-of-the-art large language models (LLMs) in a multitude of experimental setups to generate the suggestive limitations of the research papers. We conduct a comprehensive evaluation by utilizing automatic (rouge-based), human (manual expert-based), and model assistive (LLM-based) approaches.    

The key contributions of this work are: 1) To the best of our knowledge, we are the first to propose the task of Suggestive Limitation Generation (SLG) for research papers.
2) We release a SLG dataset \textbf{\textit{LimGen}}, consisting of 4068 papers and corresponding limitations.
3) We propose and experiment with several schemes to utilize LLMs for SLG.
4) We perform thorough evaluations using automated, human, and LLM-driven methodologies.\\ \\

\section{Related work}
% Scientific document understanding poses a persistent challenge, primarily attributable to its structure, diverse content modalities (such as tables and figures), and the incorporation of citations within the text. The recent emergence of large-scale scientific document summarization datasets were automatically collected from the public repositories \cite{cohan-etal-2018-discourse,delort-alfonseca-2012-dualsum}. Several works on scientific documents encompass tasks such as abstract generation \cite{kumarasinghe-de-silva-2022-automatic}, delving into the contributions outlined in a paper \cite{liu2023contributionsum,hayashi-etal-2023-whats}, scientific papers summarization \cite{cachola-etal-2020-tldr} and formulate multi-perspective summaries by leveraging reviews of the research papers \cite{cohan-etal-2022-overview-first,urlana2022ltrc}. 

% Moreover, few works delve into other forms of supervision for scientific document understanding, including citations \cite{mao-etal-2022-citesum,yasunaga2019scisummnet}, author-written highlights \cite{collins-etal-2017-supervised}, transcripts from conference presentations of the research papers \cite{lev-etal-2019-talksumm} and annotations \cite{meng-etal-2021-bringing}.  In our work, we collected the research papers and corresponding limitations from the ACL anthology. In contrast to existing works, we attempt to generate suggestive limitations of the research papers using LLMs. To the best of our knowledge, this is the first work towards generating limitations of the research papers. 

Scientific document understanding poses a persistent challenge, primarily attributable to its structure, diverse content modalities (such as tables and figures), and the incorporation of citations within the text. The recent emergence of large-scale scientific document summarization and question-answering datasets \cite{auer2023sciqa,lo2020s2orc,dasigi2021dataset} were automatically collected from the public repositories \cite{cohan-etal-2018-discourse,delort-alfonseca-2012-dualsum}. Several works on scientific documents encompass tasks such as abstract generation \cite{kumarasinghe-de-silva-2022-automatic}, delving into the contributions outlined in a paper \cite{liu2023contributionsum,hayashi-etal-2023-whats}, scientific papers summarization \cite{cachola-etal-2020-tldr} and formulate multi-perspective summaries by leveraging reviews of the research papers \cite{cohan-etal-2022-overview-first,urlana2022ltrc}. Moreover, few works delve into other forms of supervision for scientific document understanding, including citations \cite{mao-etal-2022-citesum,yasunaga2019scisummnet}, author-written highlights \cite{collins-etal-2017-supervised}, transcripts from conference presentations of the research papers \cite{lev-etal-2019-talksumm} and annotations \cite{meng-etal-2021-bringing}.  
% Close to our work, \cite{yuan2022can} attempts to generate reviews of the scientific research paper. This work formulated the scientific review generation task as an aspect-based scientific paper summarization. Further, this work utilizes the BART model with the help of the extract-and-generate paradigm (utilizing the oracle, section-based, and hybrid extraction methods). In a similar line, \cite{liu2023reviewergpt} conducted a few pilot studies to understand how large language models help in reviewing scientific papers or proposals. The study involves the identification of errors, verifying the checklists, and choosing the better papers among multiple and concludes LLMs are promising to use as a reviewing assistance. Another study \cite{xu2023exploring}, aims to generate new ideas by comprehensively examining the existing literature. This work primarily has two goals 1) Idea explorations by creating co-occurrence graphs concepts between various papers, and 2) idea verbalization, which generates fluent and reasonable texts describing the idea. 
Some recent works explore review generation \cite{yuan2022can,liu2023reviewergpt} for scientific papers by primarily leveraging publicly available data from scientific publishing platforms, which are typically written by reviewers. While peer reviews are essential for ensuring the quality and validity of research before publication, the inclusion of limitations within the paper itself offers unique advantages in terms of transparency, credibility, guidance for future research, and managing reader expectations. Limitations provide the author's perspective, providing insight into what the researchers perceive as the study's constraints (which can be agnostic to external reviewers). This internal viewpoint is different from the external critique offered by peer reviewers (which might focus heavily on the suitability of underlying research work for a particular venue), hence limitations and peer-reviews serve complementary roles in the research publication process.

We collect research papers and their corresponding author-written limitations from the ACL Anthology, specifically concentrating on generating suggestive limitations of the research papers using large language models (LLMs). To the best of our knowledge, this is the first work aimed at generating limitations for research papers.
\section{LimGen Dataset}
\subsection{Dataset collection}
We obtain the dataset from ACL Anthology\footnote{\url{https://aclanthology.org/}} website. We take advantage of the recent mandatory inclusion of the `limitations' section in the research paper for the submission of Computational Linguistics-related venues. We scrape the proceedings of EMNLP, ACL, and EACL venues of 2022 and 2023 years respectively. After obtaining the papers, the initial step involves using \texttt{scipdf\_parser}\footnote{\url{https://github.com/titipata/scipdf\_parser}} to parse the PDFs. The parser segregates the content of the paper into section-wise information. From the extracted sections, to create the `source' text for the SLG task, we discard some of the sections \textit{Abstract, Introduction, Related Works, Acknowledgements, Conclusion, Ethics Statement, Appendix, References, Limitations} and preserve the main contribution of the paper in the form of methodology, experiments, results, and discussions, etc. We use the corresponding actual `limitations' section as the reference limitations. In total, we utilize 4068 peer-reviewed short and long papers from three different ACL venues. We release the LimGen dataset under the CC BY 4.0 license. 

As an initial exploratory analysis of the proposed \textbf{LimGen} dataset, we computed several key statistics. Notably, the average length of the research papers stands at approximately 5000 tokens and 187 sentences. Whereas the limitation sections, average around 230 tokens, and 9 sentences. The longer length of the papers poses a challenge for the large language models to process the longer context length. For detailed statistics corresponding to each conference, please refer to the provided Table~\ref{tab:dataset_statistics}. 
% \usepackage{booktabs}
% \usepackage{graphicx}
\begin{table}[t]
\centering\footnotesize
\caption{LimGen Dataset Statistics}
\label{tab:dataset_statistics}
\begin{tabular}{lr|llr}
\toprule
\multicolumn{5}{c}{Number of research papers     \textbf{4068}} \\ \midrule
ACL 2022 & 1750 &  & \#Avg words per paper & 5122 \\
EMNLP 2022 & 1227 &  & \#Avg sentences per paper & 188 \\
EACL 2022 & 456 &  & \#Avg words per limitation & 230 \\
EMNLP 2023 & 635 &  & \#Avg sentences per limitation & 10 \\ \bottomrule
\end{tabular}
\end{table}

% \begin{table}[t]
% \centering\small
% \caption{LimGen Dataset Statistics}
% \begin{tabular}{lrrrrr}
% \toprule
% % \textbf{Statistic} & \textbf{Number} \\
% % \midrule
% Number of research papers & \textbf{4068} \\
% \midrule
% Number of papers per conference \\
% ACL 2022 & 1750 \\
% EMNLP 2022 & 1227 \\
% EACL 2022 & 456 \\
% EMNLP 2023 & 635 \\% Please add the following required packages to your document preamble:
% \usepackage{booktabs}
% \usepackage{graphicx}
% \begin{table}
% \centering\small
% \caption{LimGen Dataset Statistics}
% \label{tab:dataset_statistics}
% \begin{tabular}{lllll}
% \toprule
% \multicolumn{5}{l}{Number of research papers     4068} \\ \midrule
% ACL 2022 & 1750 &  & \#Avg words per paper & 5122 \\
% EMNLP 2022 & 1227 &  & \#Avg sentences per paper & 188 \\
% EACL 2022 & 456 &  & \#Avg words per limitation & 230 \\
% EMNLP 2023 & 635 &  & \#Avg sentences per limitation & 10 \\ \bottomrule
% \end{tabular}
% \end{table}

% \midrule
% Average text length & 5122.47\\
% Average text sentence count & 187.98\\
% Average Limitation length & 230.30\\
% Average Limitation sentence count & 9.56\\
% \bottomrule
% \end{tabular}
% \label{tab:dataset_statistics}
% \end{table}
% \usepackage{booktabs}
% \usepackage{graphicx}
\begin{table}
\centering\footnotesize
\setlength{\tabcolsep}{0.81ex}
\caption{Manual analysis of 60 research papers;}
\begin{tabular}{ccccccccc}
\toprule
\multicolumn{3}{c}{Relevance} & \multicolumn{3}{c}{Deduce Limitation} & \multicolumn{3}{c}{Future work or Limitation} \\ 
Yes & No & Partial & Yes & No & Partial & \multicolumn{1}{c}{\hspace{0.5cm}Yes} & \multicolumn{1}{c}{\hspace{0.3cm}No} & \multicolumn{1}{c}{Partial} \\
\cmidrule(lr){1-3}\cmidrule(lr){4-6}\cmidrule(lr){7-9}
\textbf{37} & 3 & 20 & 12 & 13 & \textbf{35} & \hspace{0.5cm}15 & \hspace{0.3cm}13 & \textbf{32} \\ 
\multicolumn{9}{c}{Limitations related to} \\ 
\cmidrule(lr){4-7}
\multicolumn{2}{c}{Methodology} & \multicolumn{3}{c}{Experimental setup} & \multicolumn{2}{c}{Dataset} & \multicolumn{2}{c}{Evaluation} \\ 
\cmidrule(lr){1-2}\cmidrule(lr){3-5}\cmidrule(lr){6-7}\cmidrule{8-9}
\multicolumn{2}{c}{\textbf{41}} & \multicolumn{3}{c}{17} & \multicolumn{2}{c}{22} & \multicolumn{2}{c}{7} \\ \bottomrule
\end{tabular}%
\label{tab:manual_analysis}
\end{table}

% % Please add the following required packages to your document preamble:
% % \usepackage{multirow}
% \begin{table*}[htb]
% \centering\small
% % \resizebox{\textwidth}{!}{%
% \caption{Manual analysis of 60 research papers}
% \setlength{\tabcolsep}{0.88ex}
% \begin{tabular}{lllllllllllll}
% \toprule
% \multicolumn{3}{c}{\textbf{Relevance}} & \multicolumn{3}{c}{\textbf{Deduce limitation}} & \multicolumn{3}{c}{\textbf{\begin{tabular}[c]{@{}l@{}}Limitation or \\ Future work\end{tabular}}} & \multicolumn{4}{c}{\textbf{Limitation related to}}                \\ 
% \cmidrule(lr){1-3}\cmidrule(lr){4-6}\cmidrule(lr){7-9}\cmidrule(lr){10-13}
% Yes      & No     & Partial     & Yes       & No       & Mix     &  \multicolumn{1}{c}{Yes}         &  \multicolumn{1}{c}{\hspace{0.4cm}No}         &  \multicolumn{1}{c}{Mix}         & Methodology & Experimental setup & Dataset & Evaluation \\ 
% \cmidrule(lr){1-3}\cmidrule(lr){4-6}\cmidrule(lr){7-9}\cmidrule(lr){10-13}
%    37   & 3  &  20  &  12  & 13  &  35  &    15  &  \multicolumn{1}{c}{\hspace{0.4cm}13}  &  \multicolumn{1}{c}{32}   &  \multicolumn{1}{c}{41} &  \multicolumn{1}{c}{17}   &  \multicolumn{1}{c}{22}  &  \multicolumn{1}{c}{7}          
%        \\ \bottomrule
% \end{tabular}
% \label{tab:manual_analysis}
% \end{table*}

\subsection{Nature of the limitations}
To understand the nature of the limitations in research papers, we conduct a manual analysis of 60 papers. We maintain diversity (short, and long papers from diverse venues) in paper selection to capture the stylistic variations of the limitations present in the research papers.  

The analysis aims to understand, 1) The relevance of the underlying limitation to the research paper, 2) whether the given limitation can be deduced by reading the paper or not?, 3) establish whether the mentioned limitation represents a real constraint or suggests potential avenues for future research, 4) classify the limitation according to its relevance to specific sections of the paper, including  \textit{Methodology}, \textit{dataset}, \textit{evaluation} or \textit{experimental setup}. We present the outcome of our manual analysis in Table~\ref{tab:manual_analysis}. We note that within our sample, numerous papers regard their future prospects as limitations. The extent of the limitation section ranges from mere sentences to substantially lengthy paragraphs. Additionally, our observations indicate that in over 50\% of the papers, discerning the stated limitation directly from the text is not straightforward. Furthermore, a significant portion of these limitations are predominantly associated with the methodology section of the paper. There are few instances, where the limitations cover more than one section of the research paper. 
% We detailed the various additional observations of the research paper limitations in Appendix~\ref{sec:limitation_examples}.   

\section{Benchmark Experiments}
\subsection{Task formulation}
This section introduces the Suggestive Limitation Generation (SLG) task formulation. 
To produce the limitations of the papers, we approach the task as a Seq2Seq problem. Precisely, we craft a model designed to intake a scientific paper \textit{R} as input and systematically produce a structured limitation block \textit{L} = \textit{l}(1:n), where \textit{l}(1:n) represents the combination of n limitations of \textit{R}, sequentially generated.
\subsection{Methodology}
This section describes the various approaches to generate the suggestive limitations. We explore two suitable text generation paradigms to generate the limitations of the research papers. Firstly, we consider the limitation generation as a summarization task and utilize the summarization-specific pre-trained models including BART\footnote{\url{https://huggingface.co/facebook/bart-large-cnn}} \cite{lewis-etal-2020-bart} and PEGASUS\footnote{\url{https://huggingface.co/google/pegasus-large}} \cite{zhang2020pegasus} to generate the suggestive limitations. Given that the objective of the SLG task surpasses the complexity of the summarization task, which typically involves limited or constrained generation entropy, the SLG task demands a higher degree of generation entropy. Unlike summarization, where the model's task is to condense information, in SLG, the model must infer and recommend limitations from the source content, drawing from its understanding during fine-tuning. Hence, the generative scheme becomes pivotal. To this end, as the second paradigm, we utilize popular Large Language Models (LLMs) namely T5\footnote{\url{https://huggingface.co/google-t5/t5-base}} \cite{raffel2020exploring}, Cerebras-GPT \cite{dey2023cerebrasgpt} and Llama 2 \cite{touvron2023llama}. To experiment with both of these paradigms, we utilize the following three schemes.  
% We intend to provide various approaches for limitation generation. Our approaches include 1) summarization-specific models, 2) utilizing generative models, and 3) investigation of the chain modeling strategies.
% This section delineates the methodologies we adopt for the task of generating suggestive limitations. Our approach is threefold: (1) application of conventional summarization models, (2) shifting to generative models, specifically Cerebras-GPT \cite{dey2023cerebrasgpt} and the state-of-the-art Llama 2 \cite{touvron2023llama}, and (3) investigation into the efficacy of chain modeling strategies. We use 80-10-10 split for train, dev, and test for the dataset. We provide a succinct description of each method alongside the experiments we employ for their setup.
\begin{figure*}[t]
    \centering
    \includegraphics[width=3.6in]{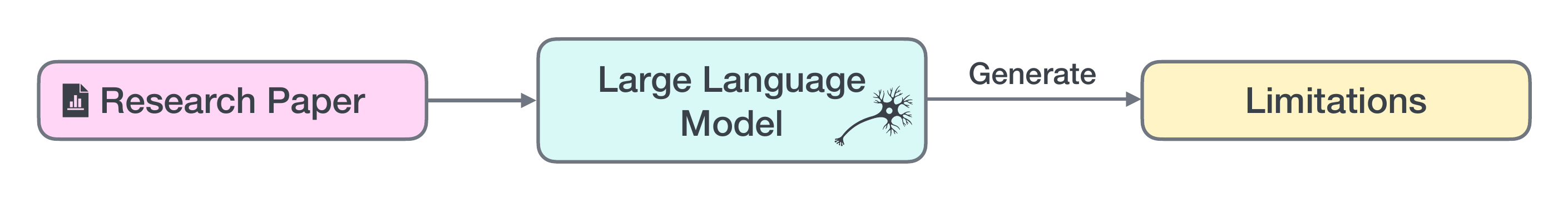}
    \caption{General architecture diagram for the suggestive limitation generation.}
    \label{fig:flow_diagram_general}
\end{figure*}
\begin{figure*}[t]
    \centering
    \includegraphics[width=4.5in]{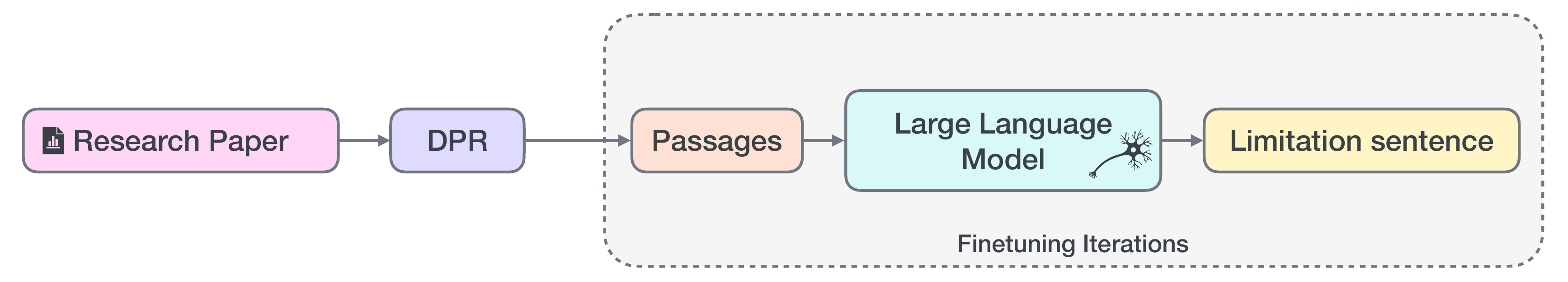}
    \caption{Architecture diagram for DPR fine-tuning.}
    \label{fig:flow_diagram_dpr_finetuning}
\end{figure*}
\begin{figure*}[t]
    \centering
    \includegraphics[width=4.9in]{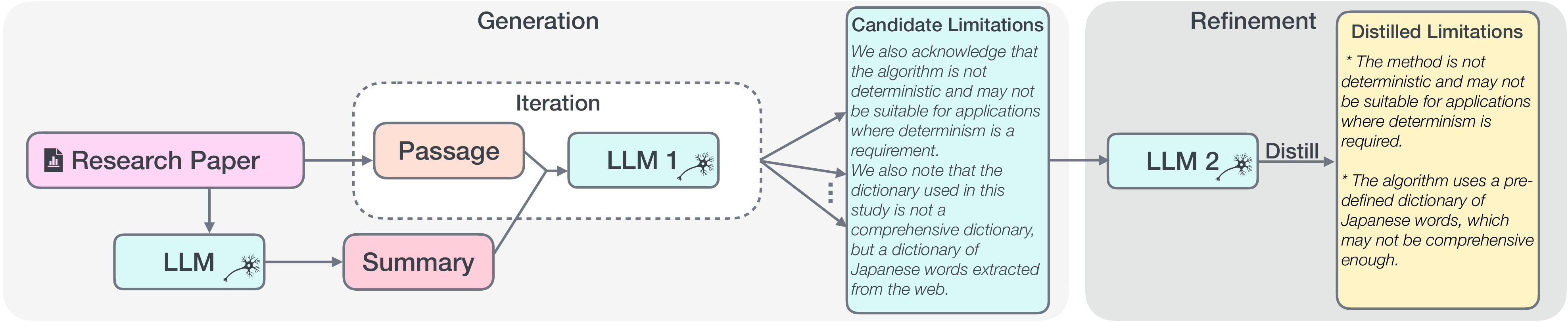}
    \caption{Architecture diagram for Chain modeling.}
    \label{fig:flow_diagram_chain_modeling}
\end{figure*}
\subsubsection{Non-truncated research paper}
In this scheme, we employ the entire research paper and its associated limitations to experiment with both summarization-specific and generative models, as illustrated in Figure~\ref{fig:flow_diagram_general}. We fine-tune the summarization models such as BART, PEGASUS and also utilize generative models including T5, Llama 2, and Cerbras GPT to perform the zero-shot prompting and fine-tuning. To experiment with the generative models in this scheme, we use the prompt depicted in Table~\ref{fig:prompt_full_finetune}. However, this approach is constrained by the models' max input token limit, resulting in a lack of comprehensive context for the research paper.
\begin{table}
\centering\footnotesize
\caption{Prompt for fine-tuning with the non-truncated research papers.}
\begin{tabular}{l} \\ \toprule
\begin{tabular}[c]{l}\textbf{Prompt:} Generate limitations or shortcomings for the following scientific paper\\ \textcolor{magenta}{\{Paper Text\}}\\ \textbf{Limitations:}\\ \textcolor{teal}{End-of-Prompt}\end{tabular} \\ \bottomrule
\end{tabular}%
\label{fig:prompt_full_finetune}
\end{table}
% Due to hardware constraints, we impose a uniform context length across all models, necessitating input truncation. Therefore, although LLMs demonstrate proficiency in the current task, their effectiveness is hindered by this truncation of the input paper. To mitigate this problem, we utilize the retrieval (DPR) approach. 

\subsubsection{Dense Passage Retrieval (DPR)}
\label{sec:dpr_method}
To address the length constraint in the Non-truncated research paper scheme, we employ the DPR approach. This approach processes only the relevant passages for each sentence present in the limitation section. To obtain the relevant passages, we utilize the following three-stage approach.
% This approach addresses the context length constraint in LLMs by processing only relevant passages for each limitation, rather than an entire truncated document, as depicted in figure \ref{fig:flow_diagram}(b). We enhance the standard dataset by including the top three relevant passages for each limitation sentence. The details of the modified dataset are below:
% \paragraph{Modified Dataset:} 
% The dataset was prepared through a multistage process:

\begin{enumerate}
    \item \textbf{Paragraph and sentence processing:} We segment the research papers into paragraphs and obtain the sentences of each paragraph by using \texttt{Spacy}\footnote{\url{https://spacy.io/api/tokenizer}}.
    % which is essential for dense passage retrieval.
    
    \item \textbf{Tokenization and paragraph management:} We tokenize these passages using BertTokenizer \cite{devlin-etal-2019-bert} to manage the length of each passage to maintain the max input token limit of 1024 for Llama 2 and 2048 for Cerebras. We merge smaller passages for optimization and split larger passages to adhere to the max input token limit. 
    
    %The input to the model consists of the paper summary, the prompt, and the input passage. We empirically find a summary of 300-350 words to be of optimal length to strike a balance between sufficient detail while mitigating issues such as redundancy and excessive focus on localized information. With the summary and the prompt, to adhere to the input token limit, the passage max length is limited to 600 tokens. Smaller passages are combined and larger passages are split into smaller segments to fit within the token limit.
    % Moreover, we merge paragraphs under 256 tokens and split those above 600 tokens. Two different passages that are split are not combined to prevent context contamination, only two full passages are combined.
    
    \item \textbf{Limitation sentence extraction and encoding:} We encode limitation sentences for each document and the passages using SentenceTransformer \cite{reimers-gurevych-2019-sentence} (all-MiniLM-L6-v2\footnote{\url{https://www.sbert.net/}}).
    % We calculate the cosine similarity between each limitation sentence and every sentence in each passage. For each limitation sentence, we include the top three passages based on these similarity scores.
    % We compute the cosine similarity between each sentence in the limitations and all sentences within each paragraph, then select only the highest similarity score for each paragraph. We discard the passages with a similarity score of less than 0.5.
    We compute the cosine similarity between each sentence in the limitations and all sentences within each passage. Further, discard the passages with a similarity score of less than 0.5. Ultimately, we identify the top three passages with the highest cosine similarity for each sentence in the limitations.
    \end{enumerate}
% \usepackage{graphicx}

% \begin{table}
% % \centering\small
% \parbox{.45\linewidth}{
% \centering\small
% \caption{Prompt utilized for both fine-tuning and generation phases in the text generation approach and the DPR approach}
% \label{fig:prompt_finetune_full}
% \begin{tabular}{l} \\ \toprule
% \begin{tabular}[c]{l}\textbf{Prompt:} Generate limitations or \\ shortcomings for the following scientific paper\\ \textcolor{magenta}{\{Paper Text\}}\\ \textbf{Limitations:}\\ \textcolor{green}{End-of-Prompt}\end{tabular} \\ \bottomrule
% \end{tabular}%
% }
% \hfill
% \parbox{.45\linewidth}{
% \centering\small
% \caption{Prompt utilized for both fine-tuning and generation phases in the text generation approach and the DPR approach.}
% \label{fig:prompt_finetune_full}
% \begin{tabular}{l} \\ \toprule
% \begin{tabular}[c]{l}\textbf{Prompt:} Generate limitations or \\ shortcomings \\ for the following passage\\ form a scientific paper\\ passage:\\ \{DPR paragraph\}\\ A brief technical summary of the\\ scientific paper for context:\\ \{Summary of the paper\}\\ Limitations:\\ End-of-Prompt\end{tabular} \\ \bottomrule
% \end{tabular}}
% \end{table}

\begin{table}[t]
\centering\footnotesize
\setlength{\tabcolsep}{0.001ex}
\caption{Prompt for DPR-based fine-tuning.}
\label{fig:prompt_finetune_full_dpr}
\begin{tabular}{l} \\ \toprule
\begin{tabular}[c]{l}\textbf{Prompt:} Generate limitations or shortcomings for the following passage from a\\ scientific paper\\ \textbf{passage:}\textcolor{magenta}{\{DPR paragraph\}}\\ A brief technical summary of the scientific paper for context:\textcolor{magenta}{\{Summary of the paper\}}\\ \textbf{Limitations:}\\ \textcolor{teal}{End-of-Prompt}\end{tabular} \\ \bottomrule
\end{tabular}
\end{table}
 We fine-tune the model by employing individual limitation sentences as target outputs, with the input comprising the top three passages retrieved through DPR (Dense Passage Retrieval). These passages are specifically obtained using the respective limitation sentence as the query. The prompt for the DPR-based fine-tuning is depicted in Table~\ref{fig:prompt_finetune_full_dpr}. The fine-tuned model is utilized to generate limitations for each passage extracted from the paper. These limitations are then compiled to form the comprehensive set of limitations associated with the research paper as shown in Figure~\ref{fig:flow_diagram_dpr_finetuning}.  Both the summarization and generative models utilize the DPR-retrieved passages for the experiments.
While generally effective, this approach may produce irrelevant limitations due to the lack of a broader context of the entire research paper. To tackle this challenge, we investigate the application of chain modeling techniques as a prospective solution.
\subsubsection{Chain Modeling}
To overcome the constraints posed by the lack of contextual information in the DPR approach, we implement the chain modeling approach inspired by \texttt{LangChain}\footnote{\url{https://github.com/LangChain-ai/LangChain}}. The chain modeling consists of two stages. In \textbf{stage 1} (\textit{limitation generation}), we generate the limitations for all passages of the research paper obtained by following the steps 1 and 2 in DPR model creation (see Section~\ref{sec:dpr_method}). However, going through individual passages, we observe that, the model lacks the comprehensive context provided by the entire research paper. To overcome this, in the prompt, we include the summary of the research paper along with each passage. The prompt to obtain the summary is mentioned in Table~\ref{fig:prompt_summary}. In \textbf{stage 2} (\textit{refinement}), we refine and distill all the generated limitations, eliminate obvious duplicates, and standardize the language as depicted in Figure~\ref{fig:flow_diagram_chain_modeling}. The prompt for both stages is shown in Table \ref{fig:prompt_iteration_reduce}.

% For this approach, we use the entire paper segmented into passages, as the primary input. This segmentation follows the same process as in the DPR approach. Next, we, step 1, generate limitations for each passage iterating through the document, and ultimately this process(chain) concludes with step 2, a final step that refines and distills all the generated limitations, eliminates obvious duplicates, and standardizes the language. We depict this in figure \ref{fig:flow_diagram}(c).
% To address the issue of lack of overall context in generating limitations we observe in the DPR approach, we pass a summary of the document along with each passage in step 1.
% The two prompts namely the iterating limitation generation prompt and the reduce or refine prompt, we use in this approach are shown in figure \ref{fig:prompt_iteration_reduce}. 
% To avoid prompt contamination, the iterating prompt is designed such that the resultant output does not contain any conversational statements, and is concise. The addition of the word ``Limitations:" at the end of the prompt as is the case in the previous approach, makes a substantial improvement in results towards this goal.
The chain modeling approach utilizes two distinct fine-tuned models. The former one (LLM1 in Figure~\ref{fig:flow_diagram_chain_modeling}) is fine-tuned on DPR passages along with the summary of the research paper as input and limitation as output. The prompt for the same is mentioned in Table~\ref{fig:prompt_finetune_full_dpr}. Whereas the latter one (LLM2 in Figure~\ref{fig:flow_diagram_chain_modeling}) is fine-tuned on full non-truncated paper as input and corresponding paper limitations as output using the prompt depicted in Table~\ref{fig:prompt_full_finetune}. This approach is indicated as \textbf{Llama2-DPR-Distilled} in Table~\ref{tab:lang_chain_resuls}. Another approach using the full non-truncated fine-tuned Llama 2 model for both LLM1 and LLM2, which is indicated as \textbf{Llama2-FT-Distilled} (FT refers to Full-text) in Table~\ref{tab:lang_chain_resuls}.
We also implement the chain modeling approach where the result of the previous iteration is passed to the current iteration with the passage. However, it is observed that when the model hallucinates or causes repetitions, all the subsequent steps are affected. For this setting, the corresponding results are reported in the table \ref{tab:lang_chain_resuls} under \textbf{Llama2-Continuous}.
\begin{table}[t]
\centering\footnotesize
\caption{Prompt for generating the summary of the research paper.}
\label{fig:prompt_summary}
\begin{tabular}{l} \\ \toprule
\begin{tabular}[c]{l}\textbf{Initial Prompt:} Write a concise technical summary of the following: \textcolor{magenta}{\{first passage\}}\\ CONCISE TECHNICAL SUMMARY: \\ \textcolor{teal}{End-of-Prompt}\\ \textbf{Iteration Prompt:} Your job is to produce a final technical summary of a research \\paper. We have provided a summary generated by you up to a certain point:\\ \textcolor{magenta}{\{summary from the previous step\}}\\ We have the opportunity to refine the existing summary (only if needed)\\ with some more context below.\\ \textcolor{magenta}{\{next passage\}}\\ Given the new context, refine the original technical summary. Keep the technical \\ summary less than 350 words. If the context isn't useful in the technical research \\context, return the original summary. Do not ask any questions in the response.\\ \textbf{Refined Technical Summary:}\\ \textcolor{teal}{End-of-Prompt}\end{tabular} \\ \bottomrule
\end{tabular}%
\end{table}
\subsection{Experimental Setup}
% Additional experimental setup details can be found in Table~\ref{tab:gen_models_parametes}.
To conduct experiments with generative models with DPR and chain modeling approaches, we use the Cerebras-GPT \cite{dey2023cerebrasgpt} 1.3B and Llama 2 \cite{touvron2023llama} 7B versions. Due to hardware constraints, we utilize the smaller versions of the LLMs and perform the fine-tuning with LoRA \cite{hu2021lora} configuration with 8-bit quantization. We utilize the XTuring code base\footnote{\url{https://github.com/stochasticai/xTuring}} for the fine-tuning. For the DPR approach, we limit the number of top passages passed to the model to 2 for Llama 2 and 3 for Cerbras-GPT. 
% The additional experimental details are mentioned in Table~\ref{tab:gen_models_parametes}. 
Moreover, we utilize four Nvidia GeForce RTX 2080 Ti GPUs (11GB). Due to increased memory requirements for finetuning Llama 2 models for chain modeling with non-truncated research papers and processing the DPR dataset with summaries as context, we temporarily upgrade to 4 NVIDIA GeForce RTX 3090 GPUs. We use an 80-10-10 split of the LimGen dataset for the creation of train, valid, and test datasets. We train Cerebras  and Llama-2 models in 3 epochs.

% For text generation experiments including DPR and Chain modelling we use the Cerebras-GPT \cite{dey2023cerebrasgpt} 1.3B version and Llama 2 \cite{touvron2023llama} 7B version. We select the above models to cover different architectures and parameter scales. Due to hardware limitations we use the 1.3B version for Cerebras-GPT and 7B version for Llama 2. We limit the context length to 1024 for Llama 2, imposing a uniform context length across all models and necessitating input truncation. Due to the same limitations, we use the LoRA \cite{hu2021lora} fine-tuning. For the generation approach and DPR approach, we use 8-bit quantization. For these approaches, the fine-tuning and text generation processes utilize the XTuring code base\footnote{\url{https://github.com/stochasticai/xTuring}}. 
% % Table \ref{tab:gen_models_parametes} lists the specific models and parameters. 
% For the DPR approach, we also limit the number of top passages passed to the model to 2 for Llama 2 and 3 for Cerbras-GPT.
% \input{table_model_parameters}

\paragraph{Specifics for Chain Modelling:}
The chain modeling approach requires more computation due to an increase in the input context length and the need for higher inference speed to process all the passages in a paper. To accommodate these requirements, we fine-tune the Llama 2 model using \texttt{Axolotl}\footnote{\url{https://github.com/OpenAccess-AI-Collective/axolotl}} with LoRA and FlashAttention \cite{dao2022flashattention}. Further, we use AWQ via \texttt{AutoAWQ}\footnote{\url{https://github.com/casper-hansen/AutoAWQ}} with a zero-point, group size of 128, 4 bits, and GEMM version on vLLM \cite{kwon2023efficient} for efficient inference. 

% \usepackage{graphicx}
\begin{table}[t]
\centering\footnotesize
\caption{Prompts for the chain modeling approach.}
\label{fig:prompt_iteration_reduce}
\begin{tabular}{l} \\ \toprule
\begin{tabular}[c]{@{}l@{}}\textbf{Iteration Prompt:} Your job is to take in a passage from a research paper\\ and a concise summary of that research and identify one or two main\\ limitations from the given passage using the summary as context.\\ \textbf{Paper Passage:} \textcolor{magenta}{\{passage\}}\\ \textbf{Paper Summary:} \textcolor{magenta}{\{summary\}}\\ \textbf{Limitations:}\\ \textcolor{teal}{End-of-Prompt}\\ \textbf{Distill Prompt:}\\ The following is set of limitations:. \textcolor{magenta}{\{list of generated limitations\}}\\ Take these and distill them into a final, consolidated list of limitations:\\ \textcolor{teal}{End-of-Prompt}\end{tabular} \\ \bottomrule
\end{tabular}
\end{table}

% For the summarization chain we generate the summary we pass in step 1 of the chain, using the model Llama 2 AWQ quantized model, from the user TheBloke on \texttt{HuggingFace}\footnote{\url{https://huggingface.co/}}.

% As these experiments require more input context length and higher inference speed as all the passages in a paper need to be processed, due to hardware constraints we leverage faster quantization techniques and high throughput inference. We fine-tune the same Llama 2 model above using \texttt{Axolotl}\footnote{\url{https://github.com/OpenAccess-AI-Collective/axolotl}} with LoRA and FlashAttention\cite{dao2022flashattention}. We use AWQ via \texttt{AutoAWQ}\footnote{\url{https://github.com/casper-hansen/AutoAWQ}} with zero-point, group size of 128, 4 bits, and GEMM version on vLLM \cite{kwon2023efficient} for efficient inference. For the summarization chain we generate the summary we pass in step 1 of the chain, using the model Llama 2 AWQ quantized model, from the user TheBloke on \texttt{HuggingFace}\footnote{\url{https://huggingface.co/}}.

\section{Experimental Results and Analysis}
\textbf{Summarization and generative approaches.} The results for experiments with summarization and generative models are listed in Table~\ref{tab:experiments_resuls}. Although the objective of generating a summary and
suggestive limitation may appear similar, our experiments reveal that models trained specifically for summarization do not effectively generate insightful limitations, often merely extracting sentences from the texts. Whereas, in the case of generative models, we observe a significant decline in output quality when excluding the `Limitations:' keyword in the prompt. In the generation approach, Llama 2 and Cerebras models demonstrate capability in limitations generation, but their effectiveness is hindered by the truncation of the input paper due to token length limitations. Moreover, they fail to generate limitations across different sections of the research paper due to the handling of limited context length. 

\noindent \textbf{Dense Passage Retrieval.} To overcome the limited context issue, we utilized the relevant passages obtained from the DPR approach to fine-tune the model. This model then iteratively takes passages and generates limitations for each passage, which are collated to get the set of limitations for a paper. This experiment's performance is generally effective when a limitation could be derived from a passage. However, despite the DPR approach being effective in highlighting relevant limitations, it inadvertently points to less pertinent, paragraph-level limitations due to the lack of overall context of the paper.

% As we move towards the DPR approach which takes the DPR retrieved passages as input, i.e. a much more relevant input, they generate much more relevant limitations. These have a much higher score than their non-DPR counterparts. Despite the DPR approach being effective in highlighting relevant limitations, it inadvertently points to less pertinent, paragraph-level limitations. 

\noindent \textbf{Chain modeling.} In this approach, we experiment with fine-tuning the models by providing the summary of the entire paper along with the passages to avoid the overall context issue. After all the limitations are generated using the input passages and summary, in the refinement stage, the duplicates are discarded. 
\begin{table*}[t]
\centering\small
\setlength{\tabcolsep}{1.1ex}
\caption{Limitation generation experimental results; R-1 refers to ROUGE-1, R-2 refers to ROUGE-2, R-L refers to ROUGE-L, BS refers to BERTScore.}
\begin{tabular}{llllllllll}
\toprule
\multicolumn{1}{c}{\multirow{2}{*}{Model}} & \multicolumn{1}{c}{\multirow{2}{*}{Approach}} & \multicolumn{4}{c|}{Without DPR data} & \multicolumn{4}{c}{With DPR data} \\  
\cmidrule{3-6} \cmidrule{7-10}  
\multicolumn{1}{c}{} & \multicolumn{1}{c}{} & \multicolumn{1}{l}{R-1} & \multicolumn{1}{l}{R-2} & \multicolumn{1}{l}{R-L} & BS & \multicolumn{1}{|l}{R-1} & \multicolumn{1}{l}{R-2} & \multicolumn{1}{l}{R-L} & BS \\ \midrule
BART-large & \multirow{3}{*}{Fine-tuning} & \multicolumn{1}{l}{30.8} & \multicolumn{1}{l}{4.5} & \multicolumn{1}{l}{15.8} & 82.8 & \multicolumn{1}{|l}{10.7} & \multicolumn{1}{l}{0.6} & \multicolumn{1}{l}{\phantom{0}8.3} & 82.8  \\  
PEGASUS-large &  & \multicolumn{1}{l}{20.2} & \multicolumn{1}{l}{3.1} & \multicolumn{1}{l}{12.7} & 82.3 & \multicolumn{1}{|l}{16.7} & \multicolumn{1}{l}{7.1} & \multicolumn{1}{l}{14.2} & 84.7 \\ 
T5-base &  & \multicolumn{1}{l}{27.7} & \multicolumn{1}{l}{4.3} & \multicolumn{1}{l}{16.4} & 83.6 & \multicolumn{1}{|l}{18.8} & \multicolumn{1}{l}{7.6} & \multicolumn{1}{l}{16.2} &  \textbf{85.9}\\ \midrule
Llama-2-7b & \multirow{2}{*}{Zero-shot} & \multicolumn{1}{l}{21.3} & \multicolumn{1}{l}{3.3} & \multicolumn{1}{l}{12.1} & 81.9  & \multicolumn{1}{|l}{16.7} & \multicolumn{1}{l}{5.2} & \multicolumn{1}{l}{8.9} &  82.4\\  
Cerebras GPT2.7B &  & \multicolumn{1}{l}{17.6} & \multicolumn{1}{l}{2.1} & \multicolumn{1}{l}{12.1} & 78.9 & \multicolumn{1}{|l}{19.8} & \multicolumn{1}{l}{4.8} & \multicolumn{1}{l}{10.3} & 80.5 \\ \midrule
Llama-2-7b & \multirow{2}{*}{Fine-tuning} & \multicolumn{1}{l}{21.4} & \multicolumn{1}{l}{3.1} & \multicolumn{1}{l}{12.7} & 81.1  & \multicolumn{1}{|l}{\textbf{34.8}} & \multicolumn{1}{l}{\textbf{11.0}} & \multicolumn{1}{l}{\textbf{17.7}} & 83.5 \\ 
Cerebras GPT2.7B &  & \multicolumn{1}{l}{16.9} & \multicolumn{1}{l}{1.9} & \multicolumn{1}{l}{12.0} & 79.1 & \multicolumn{1}{|l}{32.4} & \multicolumn{1}{l}{9.6} & \multicolumn{1}{l}{15.9} & 83.4 \\ \bottomrule
\end{tabular}
\label{tab:experiments_resuls}
\end{table*}
% \begin{table*}[ht]
% \caption{Limitation generation experimental results for chain modeling.}
% \centering \small
% \setlength{\tabcolsep}{0.85ex}
% \begin{tabular}{l|l|llll}
% \toprule
% \multicolumn{1}{c}{Chain Modeling} & \multicolumn{1}{c}{Fine-tuning} & R-1 & R-2 & R-L & BS  \\ \midrule    
% Llama-2 Continuous\\ {\footnotesize{(Iterative refinement)}} & Full Paper & \multicolumn{1}{l}{24.32} & \multicolumn{1}{l}{6.16} & \multicolumn{1}{l}{15.61} & 82.95  \\
% \midrule
% Llama2-DPR-Chain \\ (Iterative generation \\using paragraphs) & DPR dataset & \multicolumn{1}{l}{\textbf{33.52}} & \multicolumn{1}{l}{\textbf{9.90}} & \multicolumn{1}{l}{\textbf{16.30}} & 83.23  \\
% \midrule
% Llama2-DPR-Distill  & Full paper & \multicolumn{1}{l}{28.53} & \multicolumn{1}{l}{5.79} & \multicolumn{1}{l}{15.23} & \textbf{83.50}  \\ 
% \midrule
% Llama2-Full Text-Distill  & DPR dataset & \multicolumn{1}{l}{30.36} & \multicolumn{1}{l}{7.61} & \multicolumn{1}{l}{16.15} & 83.26  \\
% \bottomrule
% \end{tabular}
% \label{tab:lang_chain_resuls}
% \end{table*}

% \usepackage{graphicx}
\begin{table*}
\caption{Limitation generation experimental results for chain modeling. The model utilized for distillation/refinement consistently involves Llama2 fine-tuned on Full paper. The `Fine-tuning' column indicates the type of the dataset used for fine-tuning.}
\centering\footnotesize
\setlength{\tabcolsep}{1.1ex}
\begin{tabular}{c|c|cccc}
\toprule
Chain Modeling              & Fine-tuning & R-1  & R-2 & R-L  & BS   \\ \midrule
\begin{tabular}[c]{@{}c@{}}Llama2-Continuous\end{tabular}                     & Full Paper  & 24.3 & 6.2 & 15.6 & 82.9 \\
\begin{tabular}[c]{@{}c@{}}Llama2-DPR \end{tabular} & DPR dataset & \textbf{33.5} & \textbf{9.9} & \textbf{16.3} & 83.2 \\ \midrule
Llama2-FT-Distilled       & Full paper  & 28.5 & 5.8 & 15.2 & \textbf{83.5} \\
Llama2-DPR-Distilled & DPR dataset & 30.4 & 7.6 & 16.2 & 83.3 \\ \bottomrule
\end{tabular}%
\label{tab:lang_chain_resuls}
\end{table*}

\begin{table*}
\centering\footnotesize
\caption{Automatic evaluation using GPT-4o, utilizing xgptscore from TigerScore, shows the average penalties received for each aspect of the models. A lower score indicates better performance.}
\begin{tabular}{c|cccc}
\toprule
 & T5-base & Llama2 & Llama2-DPR & Llama2-FT-Distilled \\ \midrule
Completeness & 5.21 & 5.58 & 5.32 & \textbf{5.14} \\
Clarity & 2.79 & 2.22 & 2.34 & \textbf{1.82} \\
Relevance & \textbf{2.10} & 2.34 & 3.45 & 2.63 \\
Objectivity & \textbf{1.19} & 1.38 & 1.38 & 1.38 \\
Coherence & 2.72 & 2.39 & 2.26 & \textbf{1.68} \\
Accuracy & 2.81 & \textbf{2.57} & 3.13 & 3.72 \\ \midrule
Total & 16.82 & 16.48 & 17.88 & \textbf{16.36} \\ \bottomrule
\end{tabular}
\label{tab:gpt_tigerscore_aspects}
\end{table*}

\subsection{Evaluation}
% \usepackage{graphicx}
% \begin{table}
% \centering\small
% \caption{Prompts for GPT-4-based evaluation of the generated limitations.}
% \label{fig:prompt_evaluation}
% \begin{tabular}{l} \\ \toprule
% \begin{tabular}[c]{@{}l@{}}Evaluation Prompt: For the below sets of limitations created for the above research \\ paper, tell me if they are actually limitations and if each limitation set is proper \\ limitation of the paper even though it might not be mentioned as a limitation in \\ the original paper and rate them with respect the original limitations section in the \\ above paper and with respect to the paper itself. Also give each set a score of 1 to 5 \\ on the above qualities. 5 indicating very good limitations for the above paper. Defend \\ the score.\\ \{Limitations generated by each model\}\\ End-of-Prompt\end{tabular} \\ \bottomrule
% \end{tabular} 
% \end{table}

\begin{table}[t]
    \centering\footnotesize
    \caption{Prompts for GPT-4-based evaluation of the generated limitations.}
    \begin{tabularx}{\linewidth}{X}
        \toprule
        \textbf{Evaluation Prompt:} For the below sets of limitations created for the above research paper, tell me if they are actually limitations and if each limitation set is a proper limitation of the paper even though it might not be mentioned as a limitation in the original paper. Rate them with respect to the original limitations section in the above paper and with respect to the paper itself. Also give each set a score of 1 to 5 on the above qualities, with 5 indicating very good limitations for the above paper. Defend the score. \\ 
        \textcolor{magenta}{\{Limitations generated by each model\}} \\ 
        \textcolor{teal}{End-of-Prompt} \\ \bottomrule
    \end{tabularx}
    \label{fig:prompt_evaluation}
\end{table}

To assess the performance of the proposed models, we adopt the three types of evaluation strategies, 1) Automatic evaluation, 2) LLM-base evaluation, and 3) Human evaluation.\\
\textbf{Automatic evaluation} We conduct the automatic evaluation by using the standard evaluation metrics such as ROUGE \cite{lin-2004-rouge} and BERTScore \cite{zhang2019bertscore}. We use the RoBERTa\footnote{\url{https://huggingface.co/FacebookAI/roberta-large}} large model for the BERTScore calculation. \\
\textbf{LLM-based Evaluation} We perform the LLM-based evaluation by following two distinct approaches. The former utilizes the GPT-4 \cite{achiam2023gpt} to evaluate the quality of the generated limitations, whereas the latter is the modified version of TIGERScore\cite{jiang2023tigerscore} using GPT-4o.

\noindent \textbf{GPT-4. }In the first approach, we use the zero-shot prompting strategy to obtain the evaluation scores from GPT-4. We instructed the GPT-4 to assign a score between 1 (least) and 5 (best). As shown in Table~\ref{tab:gpt_evaluation}, we observe that the `Llama2-FT-Distilled' approach outperforms the remaining models and the T5-base obtains the lowest score. It indicates that summarization-specific models fail to generate limitations of research paper due to their inherent nature of training objective. We upload the paper as a pdf to GPT-4. The prompt used to perform the LLM-based evaluation is mentioned in Fig \ref{fig:prompt_evaluation}. We notice that GPT-4 does not thoroughly analyze the limitations. It tends to assign high scores to general limitations of the model or approach, even if they may not be accurate within the context of the provided research paper. When prompted again to verify the limitations once more, it might fail to identify incorrect limitations if they're presented in a manner that implies they pertain to any aspect discussed in the paper. When a specific limitation is singled out and prompted for re-evaluation, GPT-4 shows improved performance but struggles until the actual issue with the limitation is pointed out explicitly. 

\noindent \textbf{GPT-4o TIGERScore. }In our second approach, we employ XGPTScore, an explainable scoring method that queries GPT models using TIGERScore to interact with the GPT-4o\footnote{\url{https://openai.com/index/hello-gpt-4o/}} model. We do not use the base TIGERScore because it uses Llama 2 as the base, which cannot process entire research papers due to max token limitation. TIGERScore's evaluation method is divided into two steps: the initial freeform text assessment is followed by reformatting the output into a structured format. This evaluation scheme aims to address the decrease in output quality when GPT models are restricted to specific output formats. However, we find that this approach often fails to incorporate the first step's evaluation scores into the final formatted output. To rectify this, we manually transfer the scores from the freeform evaluations into a structured format.\\
The goal of this method is to find any issues or errors with the generated limitations and penalize each error. We evaluate the limitations based on six aspects: completeness, clarity, relevance, objectivity, coherence, and accuracy. TIGERScore assigns an aspect from these to each error and a penalty score between 0.5 and 5 for each based on its severity. The scores of these evaluations are presented in table \ref{tab:gpt_tigerscore_aspects}.\\
GPT-4o, while an improvement over its predecessor GPT-4, struggles with consistently applying penalties for completeness and sometimes accuracy in research paper assessments. It evaluates completeness by checking if a set of limitations is exhaustive and if each individual limitation is fully addressed. If a set of limitations is incomplete or an individual limitation is not thorough, GPT-4o assigns penalties. However, this evaluation is inconsistent: sometimes GPT-4o examines the overall completeness, and other times it looks at each limitation separately, rarely combining these two rules. 
% (in the prompt in figure \ref{fig:prompt_tigerscore_eval}).
This inconsistency affects the evaluation of models differently. For instance, if a model presents three limitations and none of these address a limitation GPT expects, it sometimes penalizes each of the three limitations individually rather than issuing a single penalty for the missing expected limitation. As a result, models with more limitations receive disproportionately higher penalties, despite potentially offering other valuable limitations. This skewed evaluation unfairly impacts models with more comprehensive outputs.
Despite these challenges in scoring, the Llama2-FT-Distilled model outperforms the others in overall and completeness, while the Llama2-DPR model's scores are negatively impacted by these evaluation biases.

% When prompted again to verify if the limitations are correct or not, it may overlook to catch incorrect limitations if they are framed in such a way that suggests that they are limitations to anything in the paper. When a specific limitation is singled out and prompted for re-evaluation, GPT-4 shows improved performance but struggles until the actual issue with the limitation is pointed out explicitly. 
% An example of these is included in the appendix section \ref{sec:llm-based-eval-abstract}.
% \begin{figure}
%     \centering
%     \includegraphics[width=3in]{prompt_evalutation.pdf}
%     \caption{Prompts for GPT-4-based evaluation of the generated limitations.}
%     \label{fig:prompt_evaluation}
% \end{figure}
\subsubsection{Human Evaluation} We perform the human evaluation of 50 research papers (selected randomly from the test set, in a different order) and corresponding generated limitations. Each paper is evaluated blindly by two evaluators who are industry professionals. The limitations are generated by four different systems including Llama2-DPR, Llama2, T5-base, and Llama2-FT-Distilled approaches. We asked the evaluators to rate \textit{Yes}, \textit{No}, or \textit{Partial} for each of the following questions.

\textbf{Q1.} Whether the limitation generated by the model makes sense or not?, \textbf{Q2.} Is there any overlap between the gold and generated limitation?,  \textbf{Q3.} The generated limitation is an actual limitation or not (can be a summary or prospective work), \textbf{Q4.} The generated limitation contains any hallucinations, repetitive sentences, or grammatical errors.

The results of the human evaluation are detailed in Table~\ref{tab:human_evaluation}. Our findings and manual evaluations suggest a superior performance of chain modeling using Llama2 trained on the full paper dataset for both limitation generation and reduction step (Llama2-FT-Distilled). This approach yields fewer but proper, coherent limitations, effectively reducing non-limitation and speculative content. The enhancement in the quality of generated limitations can be credited to the comprehensive summary provided to the model, enabling it to grasp the entirety of the paper's context. Also, the distillation step of the chain modeling adds coherence and structure to the results. The DPR dataset-trained model (Llama2-DPR-Distilled) closely trails behind, likely owing to its competence in generating pertinent limitations. However, it occasionally produces generic limitations rather than focused ones, such as ``The model employs Micro-F1 scores as the primary evaluation metric, yet other metrics might be more suitable depending on the particular task or application".
% or ``The model uses a simple but effective strategy to control the masking probability of each sentence dynamically, but it does not consider other strategies or techniques that may be more effective.". 
% The scores for the chain modeling approach in automatic evaluation are a bit lower when compared to the DPR approach as these are passed the entire paper (not only relevant passages) and most closely resemble real-world use cases. 
% \usepackage{booktabs}
% \usepackage{graphicx}
\begin{table}[t]
\centering\footnotesize
\caption{Automatic evaluation using GPT-4}
\setlength{\tabcolsep}{1.05ex}
\begin{tabular}{lllll}
\toprule
 & T5-base & \begin{tabular}[c]{@{}l@{}} Llama2 \end{tabular} & \begin{tabular}[c]{@{}l@{}} Llama2-DPR \end{tabular} & \begin{tabular}[c]{@{}l@{}} Llama2-FT-Distilled \end{tabular}\\
 \cmidrule{2-5}
\textbf{Score} & 2.71 &  \multicolumn{1}{c}{3.60} & \multicolumn{1}{c}{3.12} & \multicolumn{1}{c}{\textbf{4.10}}\\ \bottomrule
\end{tabular}%
\label{tab:gpt_evaluation}
\end{table}

\begin{table}
\centering\footnotesize
\caption{Human evaluation of LLMs generated limitations; For Q1-Q3, the higher values of `Yes' are preferred, whereas for Q4 the higher values of `No' are desired. All values are percentages.}
\begin{tabular}{ccccccccccccc}
\toprule
& \multicolumn{3}{c}{Llama2-DPR} & \multicolumn{3}{c}{Llama2} & \multicolumn{3}{c}{T5-base} & \multicolumn{3}{c}{Llama2-FT-Distilled} \\
\cmidrule(lr){2-4}\cmidrule(lr){5-7}\cmidrule(lr){8-10}\cmidrule(lr){11-13}
& \hspace{0.25cm}Yes\phantom{0} & No\phantom{0} & Partial & \hspace{0.25cm}Yes\phantom{0} & No\phantom{0} & Partial & \hspace{0.25cm}Yes\phantom{0} & No\phantom{0} & Partial & \hspace{0.25cm}Yes\phantom{0} & No\phantom{0} & Partial \\
\cmidrule(lr){2-4}\cmidrule(lr){5-7}\cmidrule(lr){8-10}\cmidrule(lr){11-13}
Q1 & \hspace{0.25cm}20 & 48 & 32 & 38 & 24 & 38 & 20 & 44 & 36 & \hspace{0.25cm}\textbf{70} & \phantom{0}8 & 22 \\
Q2 & \hspace{0.25cm}12 & 72 & 16 & 10 & 62 & 28 & \phantom{0}4 & 68 & 28 & \hspace{0.25cm}\textbf{28} & 30 & 42 \\
Q3 & \hspace{0.25cm}16 & 48 & 36 & 44 & 26 & 30 & 20 & 42 & 38 & \hspace{0.25cm}\textbf{62} & \phantom{0}8 & 30 \\
Q4 & \hspace{0.25cm}36 & 52 & 12 & 22 & 60 & 18 & 54 & 24 & 22 & \hspace{0.25cm}20 & \textbf{62} & 18 \\
\bottomrule
\end{tabular}
\label{tab:human_evaluation}
\end{table}
\noindent \textbf{Error analysis.} As part of human evaluation, \textbf{Q4} helps to obtain insights for the error analysis of the models. We notice that the `Llama2-FT-Distilled' model generates better limitations compared to other models. The limitations generated by `Llama2-DPR' and `T5-base' do not make sense and contain a lot of noisy information or produce the summary of the research paper rather than limitations. Moreover, the T5-base model is highly prone to directly copying several phrases from the research paper. Llama2-DPR considers local limitations, referring to limitations mentioned in a passage that refers to another paper or approach but are not the limitations of the current paper. We have also observed that more than 50\% of limitations generated from T5-base are prone to hallucination or contain repetitive phrases. Despite generating lengthy limitations, most of the `Llama2-FT-Distilled' model-generated limitations make sense.    
\section{Challenges and Future work}
In this section, we explore the key challenges and potential opportunities associated with the SLG task.\\
\textbf{Complexity of the SLG task.} As illustrated in Table \ref{tab:manual_analysis}, it is often difficult to infer the actual limitations solely from the content of the papers due to the lack of detailed context surrounding these limitations. Consequently, predicting such nuanced limitations poses a challenge for the model. Our experiments reveal that both summarization and open-text generation strategies struggle to generate these intricate limitations. Therefore, in our experiments, we place emphasis on the fine-tuning phase, with the expectation that the model will learn to discern similarities and differences across research papers, enabling it to infer nuanced limitations more effectively. Furthermore, the concept of similarity among papers can be explicitly modeled by incorporating auxiliary information, such as citations. \\
\textbf{Evaluation metrics suitability.} Lexical overlapping-based metrics such as ROUGE operate on n-gram-based matching, yet many generated limitations feature valid novel sentences and phrase formulations compared to those mentioned in the research papers. This disparity makes lexical matching-based metrics imperfect for evaluating the task of SLG. As shown in Tables~\ref{tab:experiments_resuls} and \ref{tab:lang_chain_resuls}, We find no notable discernable variation in the BERTScore values across the models. However, our human evaluation reveals considerable variations in the quality of the limitations generated by different models. Developing a tailored evaluation metric for the SLG task stands out as the most promising path forward. \\
\textbf{Multi-modal content.} We do not explore the utilization of non-textual elements such as images and tables present in the research paper to generate the limitations. Images such as architecture diagrams and plots, along with tables like result tables and ablation tables, can provide supplementary context, especially for grasping nuanced potential limitations.\\
\textbf{Coherence and relevance.} We observe that despite the apparent superior performance of DPR-based models, they fail to generate coherent limitations. The best Llama2-FT-Distilled model also generates a few speculative limitations and has difficulty in generating highly relevant limitations for every paper. \\
\textbf{Open-ended generation of LLMs.} As illustrated in Table~\ref{tab:human_evaluation}, around 20\% of the limitations generated with the aid of LLMs are susceptible to issues such as hallucination, repetitions, and grammatically incorrect sentence structures. Controlling these issues leads to the generation of more faithful limitations. \\
\textbf{Controllability.} Recent advancements in controllable generation emphasize the ability to address the specific intentions of users \cite{urlana2023controllable}. However, we note that LLMs occasionally struggle to generate specific structured limitations consistently. At times, the models produce limitations as bullet points, while in other instances, they generate them as paragraphs.
\section{Limitations}
In this work, our focus is solely on extracting limitations from only textual content within research papers. Nevertheless, incorporating non-textual elements like tables and images could enhance the generation of more contextually relevant limitations. Due to compute constraints, we experimented with the Llama 2 7B and Cerebras-GPT 1.3B models. 
% The DPR dataset retains sentences within the limitations section that may not accurately represent genuine limitations. To mitigate potential discrepancies arising from this characteristic, our methodology excludes passages exhibiting a similarity score below 0.5, but this does not constitute a flawless solution. 
The automated evaluation metrics used are insufficient for the complexity of this task, so we supplemented our evaluation process with manual evaluations. Since the LimGen dataset, sourced from the ACL anthology, it may exhibit bias towards generating limitations specific to computational linguistic aspects. As a result, its applicability to other domains like physics, chemistry, and mathematical research studies could be limited. We listed the limitations generated by our best model for this paper in Table~\ref{tab:limGen_paper_limitations}. We observe that, despite the broad coverage of the limitations, the model generates few of the sub-standard limitations.
\begin{table}[t]
    \centering\footnotesize
    \caption{Limitations generated by our best-performing model for our paper (best viewed in color). \textcolor{magenta}{Magenta} indicates the appropriate limitations, whereas \textcolor{blue}{Blue} represents sub-standard limitations.}
    \begin{tabularx}{\linewidth}{X}
        \toprule
        \textcolor{magenta}{Hallucination:} The model may generate limitations that are not present in the input text, which can lead to inaccurate or irrelevant information in the generated limitations. \textcolor{magenta}{Repetition:} The model may repeat similar phrases or sentences multiple times in the generated limitations, which can result in a lack of diversity in the generated limitations. \textcolor{magenta}{Limited training data:} The model may struggle to generate limitations that are not present in the training data, which can affect the quality of the output. \textcolor{magenta}{Inadequate evaluation metrics:} The authors do not use adequate evaluation metrics to assess the performance of their approach, which can affect the validity of their results. \textcolor{blue}{Lack of consideration of the research question:} The authors do not consider the research question when generating limitations, which can lead to inaccurate or irrelevant limitations. \textcolor{blue}{Lack of consideration of the methodology:} The authors do not consider the methodology used in the research paper when generating limitations, which can lead to inaccurate or irrelevant limitations.\\
        \bottomrule
    \end{tabularx}%
    \label{tab:limGen_paper_limitations}
\end{table}

\section{Conclusions}
In this paper, we introduce the novel task of Suggestive Limitation Generation (SLG) for research papers, aiming to provide reviewers with potential limitations of the underlying work, thereby assisting in the review process. We compile a dataset of 4068 research papers and corresponding limitations from the ACL anthology. We propose an LLM-based baseline for SLG tasks and conduct several ablation studies. Subsequently, we perform a thorough evaluation of the proposed models with automatic, LLM-based, and human evaluation schemes. Moving forward, our plans involve incorporating images and tabular content in addition to text for the SLG task.

\section{Ethics Statement}
In the creation of the LimGen dataset, we did not collect any personal data. Instead, we rely on the publicly accessible dataset from the ACL anthology. We release the LimGen dataset under the Creative Commons Attribution 4.0 International license (same as the original license of ACL anthology). We release the code under the MIT license.

\begin{thebibliography}{8}
\providecommand{\url}[1]{\texttt{#1}}
\providecommand{\urlprefix}{URL }
\providecommand{\doi}[1]{https://doi.org/#1}

\bibitem{achiam2023gpt}
Achiam, J., Adler, S., Agarwal, S., Ahmad, L., Akkaya, I., Aleman, F.L., Almeida, D., Altenschmidt, J., Altman, S., Anadkat, S., et~al.: Gpt-4 technical report. arXiv preprint arXiv:2303.08774  (2023)

\bibitem{auer2023sciqa}
Auer, S., Barone, D.A., Bartz, C., Cortes, E.G., Jaradeh, M.Y., Karras, O., Koubarakis, M., Mouromtsev, D., Pliukhin, D., Radyush, D., et~al.: The sciqa scientific question answering benchmark for scholarly knowledge. Scientific Reports  \textbf{13}(1), ~7240 (2023)

\bibitem{cachola-etal-2020-tldr}
Cachola, I., Lo, K., Cohan, A., Weld, D.S.: Tldr: Extreme summarization of scientific documents. In: Findings of the Association for Computational Linguistics: EMNLP 2020. pp. 4766--4777 (2020)

\bibitem{cohan-etal-2018-discourse}
Cohan, A., Dernoncourt, F., Kim, D.S., Bui, T., Kim, S., Chang, W., Goharian, N.: A discourse-aware attention model for abstractive summarization of long documents. In: Proceedings of the 2018 Conference of the North American Chapter of the Association for Computational Linguistics: Human Language Technologies, Volume 2 (Short Papers). pp. 615--621 (2018)

\bibitem{cohan-etal-2022-overview-first}
Cohan, A., Feigenblat, G., Ghosal, T., Shmueli-Scheuer, M.: Overview of the first shared task on multi perspective scientific document summarization (mup). In: Proceedings of the Third Workshop on Scholarly Document Processing. pp. 263--267 (2022)

\bibitem{collins-etal-2017-supervised}
Collins, E., Augenstein, I., Riedel, S.: A supervised approach to extractive summarisation of scientific papers. In: Proceedings of the 21st Conference on Computational Natural Language Learning (CoNLL 2017). pp. 195--205 (2017)

\bibitem{dao2022flashattention}
Dao, T., Fu, D., Ermon, S., Rudra, A., R{\'e}, C.: Flashattention: Fast and memory-efficient exact attention with io-awareness. Advances in Neural Information Processing Systems  \textbf{35},  16344--16359 (2022)

\bibitem{dasigi2021dataset}
Dasigi, P., Lo, K., Beltagy, I., Cohan, A., Smith, N.A., Gardner, M.: A dataset of information-seeking questions and answers anchored in research papers. In: Proceedings of the 2021 Conference of the North American Chapter of the Association for Computational Linguistics: Human Language Technologies. pp. 4599--4610 (2021)

\bibitem{delort-alfonseca-2012-dualsum}
Delort, J.Y., Alfonseca, E.: Dualsum: a topic-model based approach for update summarization. In: Proceedings of the 13th Conference of the European Chapter of the Association for Computational Linguistics. pp. 214--223 (2012)

\bibitem{devlin-etal-2019-bert}
Devlin, J., Chang, M.W., Lee, K., Toutanova, K.: Bert: Pre-training of deep bidirectional transformers for language understanding. In: Proceedings of the 2019 Conference of the North American Chapter of the Association for Computational Linguistics: Human Language Technologies, Volume 1 (Long and Short Papers). pp. 4171--4186 (2019)

\bibitem{dey2023cerebrasgpt}
Dey, N., Gosal, G., Zhiming, Chen, Khachane, H., Marshall, W., Pathria, R., Tom, M., Hestness, J.: Cerebras-gpt: Open compute-optimal language models trained on the cerebras wafer-scale cluster (2023)

\bibitem{hayashi-etal-2023-whats}
Hayashi, H., Kry{\'s}ci{\'n}ski, W., McCann, B., Rajani, N., Xiong, C.: What’s new? summarizing contributions in scientific literature. In: Proceedings of the 17th Conference of the European Chapter of the Association for Computational Linguistics. pp. 1019--1031 (2023)

\bibitem{hu2021lora}
Hu, E.J., Wallis, P., Allen-Zhu, Z., Li, Y., Wang, S., Wang, L., Chen, W., et~al.: Lora: Low-rank adaptation of large language models. In: International Conference on Learning Representations (2021)

\bibitem{jiang2023tigerscore}
Jiang, D., Li, Y., Zhang, G., Huang, W., Lin, B.Y., Chen, W.: Tigerscore: Towards building explainable metric for all text generation tasks. arXiv preprint arXiv:2310.00752  (2023)

\bibitem{kumarasinghe-de-silva-2022-automatic}
Kumarasinghe, D., de~Silva, N.: Automatic generation of abstracts for research papers. In: Proceedings of the 34th Conference on Computational Linguistics and Speech Processing (ROCLING 2022). pp. 221--229 (2022)

\bibitem{kwon2023efficient}
Kwon, W., Li, Z., Zhuang, S., Sheng, Y., Zheng, L., Yu, C.H., Gonzalez, J., Zhang, H., Stoica, I.: Efficient memory management for large language model serving with pagedattention. In: Proceedings of the 29th Symposium on Operating Systems Principles. pp. 611--626 (2023)

\bibitem{lev-etal-2019-talksumm}
Lev, G., Shmueli-Scheuer, M., Herzig, J., Jerbi, A., Konopnicki, D.: Talksumm: A dataset and scalable annotation method for scientific paper summarization based on conference talks. In: Proceedings of the 57th Annual Meeting of the Association for Computational Linguistics. pp. 2125--2131 (2019)

\bibitem{lewis-etal-2020-bart}
Lewis, M., Liu, Y., Goyal, N., Ghazvininejad, M., Mohamed, A., Levy, O., Stoyanov, V., Zettlemoyer, L.: Bart: Denoising sequence-to-sequence pre-training for natural language generation, translation, and comprehension. In: Proceedings of the 58th Annual Meeting of the Association for Computational Linguistics. pp. 7871--7880 (2020)

\bibitem{lin-2004-rouge}
Lin, C.Y.: {ROUGE}: A package for automatic evaluation of summaries. In: Text Summarization Branches Out. pp. 74--81. Association for Computational Linguistics, Barcelona, Spain (Jul 2004)

\bibitem{liu2023contributionsum}
Liu, M.H., Yen, A.Z., Huang, H.H., Chen, H.H.: Contributionsum: Generating disentangled contributions for scientific papers. In: Proceedings of the 32nd ACM International Conference on Information and Knowledge Management. pp. 5351--5355 (2023)

\bibitem{liu2023reviewergpt}
Liu, R., Shah, N.B.: Reviewergpt? an exploratory study on using large language models for paper reviewing. arXiv preprint arXiv:2306.00622  (2023)

\bibitem{lo2020s2orc}
Lo, K., Wang, L.L., Neumann, M., Kinney, R., Weld, D.S.: S2orc: The semantic scholar open research corpus. In: Proceedings of the 58th Annual Meeting of the Association for Computational Linguistics. pp. 4969--4983 (2020)

\bibitem{mao-etal-2022-citesum}
Mao, Y., Zhong, M., Han, J.: {C}ite{S}um: Citation text-guided scientific extreme summarization and domain adaptation with limited supervision. In: Goldberg, Y., Kozareva, Z., Zhang, Y. (eds.) Proceedings of the 2022 Conference on Empirical Methods in Natural Language Processing. pp. 10922--10935. Association for Computational Linguistics, Abu Dhabi, United Arab Emirates (Dec 2022)

\bibitem{meng-etal-2021-bringing}
Meng, R., Thaker, K., Zhang, L., Dong, Y., Yuan, X., Wang, T., He, D.: Bringing structure into summaries: a faceted summarization dataset for long scientific documents. In: Proceedings of the 59th Annual Meeting of the Association for Computational Linguistics and the 11th International Joint Conference on Natural Language Processing (Volume 2: Short Papers). pp. 1080--1089 (2021)

\bibitem{raffel2020exploring}
Raffel, C., Shazeer, N., Roberts, A., Lee, K., Narang, S., Matena, M., Zhou, Y., Li, W., Liu, P.J.: Exploring the limits of transfer learning with a unified text-to-text transformer. The Journal of Machine Learning Research  \textbf{21}(1),  5485--5551 (2020)

\bibitem{reimers-gurevych-2019-sentence}
Reimers, N., Gurevych, I.: Sentence-bert: Sentence embeddings using siamese bert-networks. In: Proceedings of the 2019 Conference on Empirical Methods in Natural Language Processing and the 9th International Joint Conference on Natural Language Processing (EMNLP-IJCNLP). pp. 3982--3992 (2019)

\bibitem{touvron2023llama}
Touvron, H., Martin, L., Stone, K., Albert, P., Almahairi, A., Babaei, Y., Bashlykov, N., Batra, S., Bhargava, P., Bhosale, S., et~al.: Llama 2: Open foundation and fine-tuned chat models. arXiv preprint arXiv:2307.09288  (2023)

\bibitem{urlana2023controllable}
Urlana, A., Mishra, P., Roy, T., Mishra, R.: Controllable text summarization: Unraveling challenges, approaches, and prospects--a survey. arXiv preprint arXiv:2311.09212  (2023)

\bibitem{urlana2022ltrc}
Urlana, A., Surange, N., Shrivastava, M.: Ltrc@ mup 2022: Multi-perspective scientific document summarization using pre-trained generation models. In: Proceedings of the Third Workshop on Scholarly Document Processing. pp. 279--284 (2022)

\bibitem{yasunaga2019scisummnet}
Yasunaga, M., Kasai, J., Zhang, R., Fabbri, A.R., Li, I., Friedman, D., Radev, D.R.: Scisummnet: A large annotated corpus and content-impact models for scientific paper summarization with citation networks. In: Proceedings of the AAAI conference on artificial intelligence. vol.~33, pp. 7386--7393 (2019)

\bibitem{yuan2022can}
Yuan, W., Liu, P., Neubig, G.: Can we automate scientific reviewing? Journal of Artificial Intelligence Research  \textbf{75},  171--212 (2022)

\bibitem{zhang2020pegasus}
Zhang, J., Zhao, Y., Saleh, M., Liu, P.: Pegasus: Pre-training with extracted gap-sentences for abstractive summarization. In: International Conference on Machine Learning. pp. 11328--11339. PMLR (2020)

\bibitem{zhang2019bertscore}
Zhang, T., Kishore, V., Wu, F., Weinberger, K.Q., Artzi, Y.: Bertscore: Evaluating text generation with bert. arXiv preprint arXiv:1904.09675  (2019)

\end{thebibliography}
\end{document}